\documentclass{article}

\usepackage{microtype}
\usepackage{graphicx}
\usepackage{subfigure}
\usepackage{booktabs} % for professional tables

\usepackage{hyperref}

\usepackage[accepted]{icml2024}

\usepackage{amsmath}
\usepackage{amssymb}
\usepackage{mathtools}
\usepackage{amsthm}

\usepackage{algorithm}
\usepackage{caption}
\usepackage{colortbl}
\usepackage{makecell, multirow}
\usepackage{xcolor}

\usepackage[capitalize,noabbrev]{cleveref}

\theoremstyle{plain}

\theoremstyle{definition}

\theoremstyle{remark}

\usepackage[textsize=tiny]{todonotes}

\icmltitlerunning{
Visually Dehallucinative Instruction Generation:
Know What You Don't Know
}

\begin{document}

\twocolumn[
\icmltitle{Visually Dehallucinative Instruction Generation: \\
    Know What You Don't Know}




\begin{icmlauthorlist}
\icmlauthor{Sungguk Cha}{nc}
\icmlauthor{Jusung Lee}{nc}
\icmlauthor{Younghyun Lee}{nc}
\icmlauthor{Cheoljong Yang}{nc}
\end{icmlauthorlist}

\icmlaffiliation{nc}{Multimodal AI Lab., NC Research, NCSOFT Corporation}

\icmlcorrespondingauthor{Sungguk Cha}{sungguk@ncsoft.com}
\icmlcorrespondingauthor{Cheoljong Yang}{cjyang@ncsoft.com}

\icmlkeywords{visual hallucination, dehallucination}

\vskip 0.3in
]



\printAffiliationsAndNotice{}  % leave blank if no need to mention equal contribution

\newcommand{\idk}{IDK-Instructions}
\newcommand{\idkb}{IDK-Instructions }
\newcommand{\hal}{IK hallucination}
\newcommand{\halb}{IK hallucination }
\newcommand{\vqa}{VQAv2-IDK}
\newcommand{\vqab}{VQAv2-IDK }
\newcommand{\eg}{\textit{e}.\textit{g}., }
\newcommand{\ie}{\textit{i}.\textit{e}., }

\begin{abstract}
\textit{"When did the emperor Napoleon invented iPhone?"}
Such hallucination-inducing question is well known challenge in generative language modeling. 
In this study, we present an innovative concept of visual hallucination, referred to as \textit{"I Know (IK)" hallucination}, to address scenarios where "I Don't Know" is the desired response.
To effectively tackle this issue, we propose the \vqab benchmark, the subset of VQAv2 comprising unanswerable image-question pairs as determined by human annotators. 
Stepping further, we present the visually dehallucinative instruction generation method for IK hallucination and introduce the \idkb visual instruction database.
Our experiments show that current methods struggle with IK hallucination. 
Yet, our approach effectively reduces these hallucinations, proving its versatility across different frameworks and datasets.
\end{abstract}

    \begin{figure}[!ht]
\centering
\centerline{\includegraphics[width=1.0\linewidth]{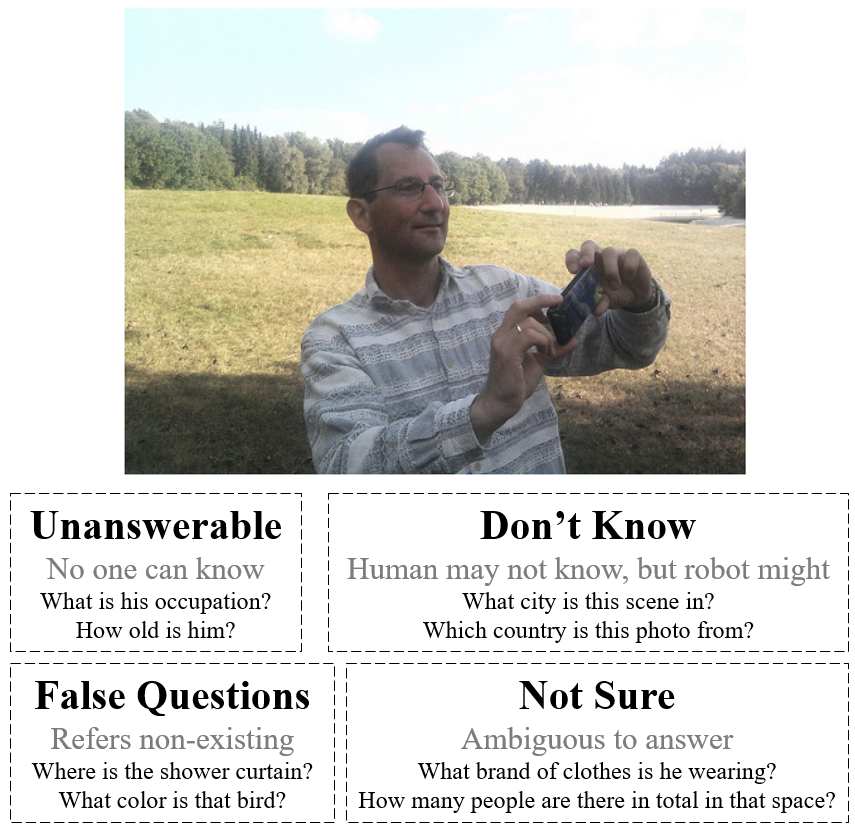}}
\caption{
IK hallucination categories with examples.
}
\label{fig:category}
\end{figure}
\section{Introduction}
% Introduce problem
The recent advancements in large multi-modal models (LMMs) have manifested in our daily lives through the introduction of chatbot assistants such as GPT-4V~\cite{openai23gpt4v} and Gemini~\cite{geminiteam2023gemini}. 
Amidst these advancements, visual hallucination has risen as a significant concern and is considered a crucial focal point.
Nevertheless, the current landscape of visual hallucination is limited to the binary determination of object presence or absence~\cite{rohrbach2018object, li2023evaluating}.

% This paper
%% introduce IDK
Expanding on this context, our research introduces a novel visual hallucination concept termed I Know (IK) hallucination, particularly addressing situations where providing definitive answers is challenging (see Fig.~\ref{fig:category}). 
In the context of Visual Question Answering (VQA), where models typically learn from true positives only, language modeling approaches strive to generate responses even in hallucination-inducing scenarios.
Our objective is to address \hal, aiming to avoid the generation of arbitrary statements by instructing the model to respond with "I don't know" when uncertain.

\begin{figure}[ht!]
\centering
\begin{minipage}[]{\linewidth}
  \centering
  \centerline{\includegraphics[width=0.75\linewidth]{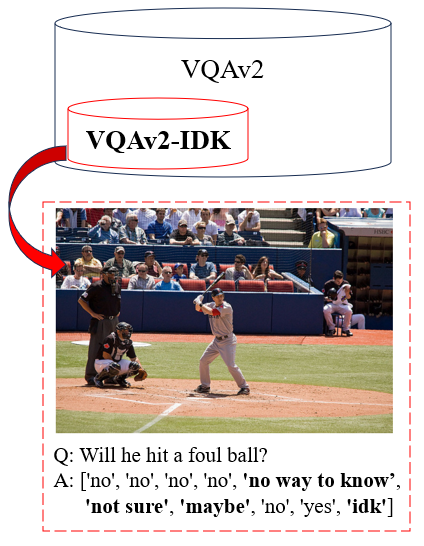}}
%  \vspace{2.0cm}
  \centerline{(a) Collect ‘hallucination-inducing’ VQAs}\medskip
\end{minipage}
\begin{minipage}[]{\linewidth}
    \centering
    \begin{tabular}{|p{6.5cm}|}
    \toprule
    \multicolumn{1}{|c|}{\small \textbf{Few-shot prompts}} \\
    \midrule
    {\small \# I Don’t Know Aware Visual Instruction Generation} \\
    {\small \#\# Task Description} \\
    {\small Carefully filtered VQA will be given, which is composed of a question and answers without an image and contains questions those are ambiguous to answer. The task is to generate proper answer.} \\
    {\small \#\# Examples} \\
    {\small \#\#\# VQA Q: Is the bird female? A: ['yes', 'yes', 'yes', "i don't know", 'no', 'yes', 'maybe', 'yes', 'yes', 'no’]} \\
    {\small \#\#\# Answer} \\
    {\small I don't know if the bird is female. It can be both male and female.} \\
    {\small ... $\langle skip \rangle$ } \\
    \midrule
    \multicolumn{1}{|c|}{\small \textbf{VQAv2-IDK sample}} \\
    \midrule
    {\small Q: Will he hit a foul ball?} \\
    {\small A: ['no', 'no', 'no', 'no', 'no way to know’, 'not sure', 'maybe', 'no', 'yes', 'idk']} \\
    \bottomrule
    \end{tabular}
    \centerline{\LARGE \rotatebox[origin=c]{270}{\textcolor{orange}{$\Rightarrow$}}}
    \begin{tabular}{|c|}
        \hline
        {Language Model} \\
        \hline
    \end{tabular}
    \centerline{\LARGE \rotatebox[origin=c]{270}{\textcolor{orange}{$\Rightarrow$}}}
    \vspace{3pt} \centerline{\textbf{IDK-Instructions}}
    \arrayrulecolor{red}
    \vspace{10pt}
    \begin{tabular}{|c|}
        \hline
        \textcolor{red}{A: It is unclear whether he will hit a foul ball or not.} \\
        \hline
    \end{tabular}
    \arrayrulecolor{black}
    \centerline{(b) IDK instruction generation}
\end{minipage}
\caption{
IK hallucination aware visually dehallucinative instruction generation overview. 
}
\label{fig:approach}
\end{figure}

%% VQAv2-IDK
To explore and evaluate \halb concept, we propose a novel benchmark named \vqa, which is the subset of the VQAv2~\cite{balanced_vqa_v2}, comprising of hallucination-inducing VQAs (see Fig.~\ref{fig:approach} (a)).
During our investigation, we have observed the presence of hallucination-inducing image-question pairs across various VQA datasets such as VQAv2 and OKVQA~\cite{okvqa}. 
These pairs come with the corresponding human responses, \eg "I don't know" or "unanswerable."
\vqab consists of such deliberately collected human annotations in which "I don't know" becomes the desired answer. 

%% IDK-instructions
As a remedy, we introduce a visually dehallucinative instruction generation method designed to mitigate the effects of \hal, presenting the resulting \idkb (refer to Fig.~\ref{fig:approach} (b)).
Drawing inspiration from \cite{cha2024visually}, which crafts instructions exclusively based on provided information, our framework produces proper responses to the curated hallucination-inducing image-question pairs. 
The generation process involves employing language models with the carefully constructed few-shot prompt. 
As a result, we present the IK-dehallucinative visual instruction dataset, dubbed \idk.

% Experiments
In our experiments, we confirm the vulnerability of existing works to \hal.
Further, our experiments show \idkb effectively mitigates \hal.
It shows the versatility seamlessly incorporating with any frameworks and any dataset combinations, while preserving visual recognition performances. 

% Summary
In summary, our contributions are:
\begin{itemize}
    \item Expanding the assessable scope of visual hallucination, we introduce a novel concept of visual hallucination, termed \textbf{\textit{IK hallucination}}, wherein "I don't know" is the desired answer. 
    Also we propose IK hallucination benchmark \textbf{\textit{VQAv2-IDK}}, the subset of VQAv2, comprising of image-question pairs that intrigue hallucination.
    \item We present visually dehallucinative instruction generation method for IK hallucination. 
    Our generated visual instruction, \textbf{\textit{\idkb}}, shows the efficacy on mitigating IK hallucination and the versatility by incorporating with any frameworks and any datasets. 
    \item We publicly release IK benchmark (VQAv2-IDK), and the generated visually dehallucinative instruction dataset (\idk). \footnote{https://github.com/ncsoft/idk}
\end{itemize}
\section{Background}
\textbf{Hallucination}
Hallucinations from generative language models, involving the generation of incorrect or improbable information, are regarded as pivotal concerns for reliability and have been extensively documented~\cite{openai2023gpt4, openai23gpt4v, geminiteam2023gemini}.
Visual hallucination is precisely defined by the generation of deceptive or inaccurate visual content, presenting significant challenges in maintaining the fidelity and accuracy of the model's output.
In recent endeavors to tackle this issue, there have been efforts to detect~\cite{gunjal2023detecting} or quantitatively evaluate visual hallucination~\cite{rohrbach2018object, li2023evaluating, zhai2023halle}. %, along with mitigating approaches~\cite{}.
Our work introduces a novel approach to evaluating visual hallucination, along with a visually dehallucinative instruction generation method.

\textbf{Visual Hallucination Evaluation}
Despite recent explorations in visual hallucination primarily focusing on recognitive aspects, such as presence/absence~\cite{rohrbach2018object, li2023evaluating} and grounding~\cite{zhai2023halle}, there remains a notable gap in practice when it comes to hallucination intriguing cases, for example, unanswerable questions~\cite{rajpurkar2018know}.
The following discussions aim to address this void by delving into the nuanced realm of hallucination, particularly focusing on intriguing cases, such as unanswerable questions, that have been relatively underexplored in the current landscape of visual hallucination evaluation.

\textbf{Visual Instruction Generation}
While instruction tuning~\cite{ouyang2022training} enhances pretrained language models for conversational interactions, expanding their proficiency to encompass vision-language modalities~\cite{instructblip, liu2023llava}, the acquisition of instructions remains a significant challenge. 
Notably, \cite{instructblip, liu2023llava} suggested the conversion of existing databases (e.g., image-caption, detection, and segmentation) into VQA format to enhance visual recognition capabilities. 
Additionally, \cite{cha2024visually, liu2023llava} proposed instruction generation methods utilizing language models. 
Particularly, \cite{cha2024visually} introduced a visually dehallucinative instruction generation method designed to mitigate the generation of non-factual contents. 
Building upon this, our work contributes by introducing a method for generating instructions tailored to address various unanswerable questions, along with the generated instructions.
    \begin{figure}[!t]
\begin{center}
\begin{tabular}{p{1.5cm}p{5.7cm}}
\toprule
\multicolumn{2}{c}{\textbf{\idkb and Its VQAv2 Sources}} \\
\midrule
& \includegraphics[width=0.8\linewidth]{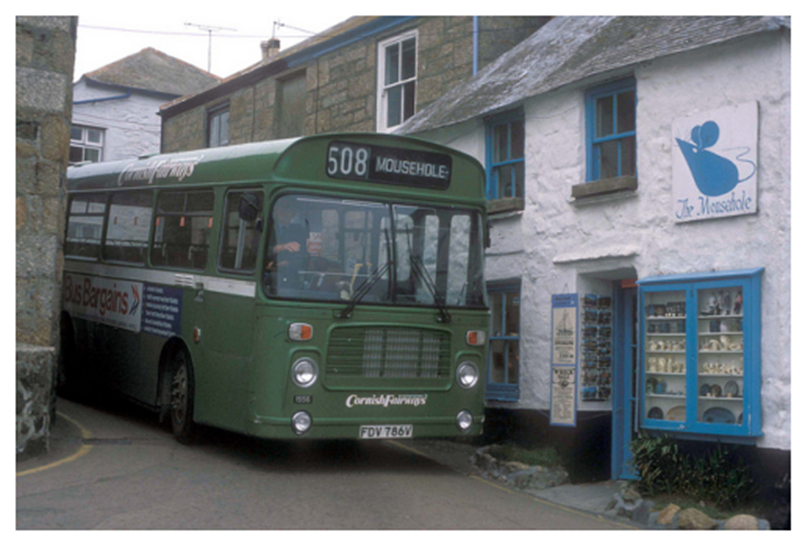} \\
Question & How quickly can I get to the food market? Is it accessible? \\
VQAv2 & ['5 minutes', '1 min', 'fast yes', \textbf{'not sure yes'}, '5 min yes', 'irrelevant', 'timely', 'yes', '5 minutes yes', 'not possible’] \\
\idk & I am \textbf{not sure} how quickly you can get to the food market. It may take about 5 minutes or 1 minute. And I am also not sure if it is accessible. \\
\midrule
& \includegraphics[width=0.8\linewidth]{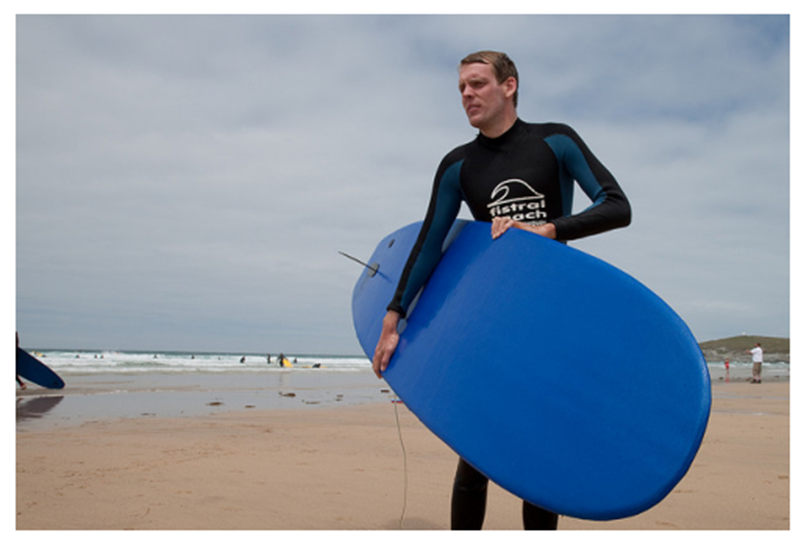} \\
Question & How has this photo been edited? \\
VQAv2 & [\textbf{'unknown'}, 'no', 'cropped', 'not edited', 'close up’, 
        'no', 'no', 'no', 'yes', 'poorly’] \\
\idk & \textbf{I don't know} how this photo has been edited. It can be cropped or not edited at all. \\
\bottomrule
\end{tabular}
\end{center}
\caption{VQAv2 questions, answers and the generated \idkb. IDK keywords are marked bold.}
\label{fig:idk}
\end{figure}  
    \begin{table}[t]
\caption{
    Occurrences of IDK Keywords in the VQAv2 train and validation splits.
    It represents how many times each keyword appears in the human annotation. 
}
\label{table:vqav2idk}
\begin{center}
\begin{small}
\begin{sc}
\begin{tabular}{lcc}
\toprule
Keyword & Train & Val \\
\midrule
none & 10,859 & 4,293 \\
unknown & 4,587 & 2,184 \\
don't know & 3,007 & 1,424 \\
not sure & 2,143 & 1,233 \\
unanswerable & 74 & 366 \\
uncertain & 38 & 18 \\
bad question & 8 & 5 \\
depends & 24 & 4 \\
ambiguous & 1 & 1 \\
%\midrule
%Total QA & 13,807 & 6,624 \\
\bottomrule
\end{tabular}
\end{sc}
\end{small}
\end{center}
\end{table}
    \begin{table}[t]
\caption{Distribution of IDK Categories in VQAv2-IDK train and validation sets. The digits refer the number of QA instances containing IDK keyword as an answer.}
\label{table:category}
\begin{center}
\begin{small}
\begin{sc}
\begin{tabular}{lcc}
\toprule
Category & Train QA & Val QA \\
\midrule
Unanswerable & 4,352 & 2,357 \\
False Questions & 5,671 & 2,289 \\
Don't Know & 2,870 & 1,343 \\
Not Sure & 2,143 & 1,220 \\
\bottomrule
\end{tabular}
\end{sc}
\end{small}
\end{center}
\end{table}
    \begin{algorithm}[t]
    \caption{Filtering VQAv2 with IDK keywords}
    \label{alg:vqav2idk}
    
    \begin{algorithmic}[1]
        \STATE \textbf{Input:} data $VQAv2$, keywords $Keywords$
        \STATE Initialize $VQAv2$-$IDK$ as empty list
        
        \FOR{$VQA(I, Q, As)$ in $VQAv2$}
            \STATE Initialize $added = \text{False}$
            
            \FOR{$A$ in $As$}
                \FOR{$Keyword$ in $Keywords$}
                    \IF{$A$ contains $Keyword$}
                        \STATE $VQAv2$-$IDK$ += [$VQA$]
                        \STATE $added = \text{True}$
                        \STATE \textbf{break}
                    \ENDIF
                \ENDFOR
                
                \IF{$added$}
                    \STATE \textbf{break}
                \ENDIF
            \ENDFOR
        \ENDFOR
    \end{algorithmic}
\end{algorithm}
\section{IK Hallucination Benchmark: VQAv2-IDK}
\label{section:vqaidk}

\subsection{Defining IK Hallucination}
\label{section:definition}
% 그래서 그 문제를 풀려다가 이걸 하게됐다
In this work, we define "IK hallucinated" as responding with anything other than "I don't know" to a question where "I don't know" should have been the appropriate answer.
The VQA datasets that the most LMMs leverage for visual instruction tuning consists of human annotations~\cite{antol2015vqa, balanced_vqa_v2, okvqa}, \ie the question and the answers are human crafted. 
Despite their efforts to ensure "question makes sense", there still remains questions that do not make sense, so is unanswerable (see Fig. \ref{fig:category}, \ref{fig:approach}.(a) and \ref{fig:idk}). 

\subsection{Deriving VQAv2-IDK from VQAv2}
We introduce VQAv2-IDK, the subset of VQAv2 dataset, in which human annotators have annotated them with IDK keywords. 
It serves as a VQA benchmark where the anticipated answer is "I don't know."
We investigated common answer keywords for the hallucination-inducing questions and collected VQA instances containing the keyword in the answer list (see Algorithm~\ref{alg:vqav2idk}). 
Table~\ref{table:vqav2idk} shows the occurrences of IDK keywords in the VQAv2 dataset. 
The training set and validation set of VQAv2-IDK consist of $13,807$ and $6,624$ image-question pairs respectively.

\subsection{IDK Categorization}
\label{section:category}
We classified the identified IDK keywords into four distinct categories (see also Table~\ref{table:category}):
\begin{itemize}
    \item Unanswerable (nothing can know): unknown, unanswerable, bad question, depends, ambiguous
    \item False questions (referring non-existing): none
    \item Don't know (human may not know but robot might): don't know
    \item Not sure (ambiguous or not enough to answer): not sure, uncertain
\end{itemize}

Figure~\ref{fig:category} illustrates the IDK categories.
The "Unanswerable" category includes cases where the question is inherently unanswerable, \ie neither human nor machine can know.
The "False questions" category contains questions that refer to non-existing objects, which directly intrigues hallucination.
The "Don't know" category includes questions where a human may not know the answer but a robot might.
The "Not sure" category covers unclear cases where the image \textit{I} does not provide enough information for identification.

We categorize questions into "unanswerable" and "don't know" to distinguish between those that are universally impossible to answer and those that may be unanswered by humans but could potentially be answered by a robot, allowing for differentiation based on the scope of knowledge.
In practical terms, if a model excels at answering questions categorized as "unanswerable" but performs poorly on questions labeled as "don't know," it suggests the model is resilient to "I know" hallucination and may possess knowledge beyond that of humans.
    \begin{figure*}[ht]
\begin{center}
\begin{tabular}{p{1cm}p{6cm}p{6cm}}
\toprule
    \multicolumn{3}{c}{\textbf{Visual Input Examples: Hallucination-Inducing Questions}} \\
    \midrule
    & \includegraphics[width=\linewidth]{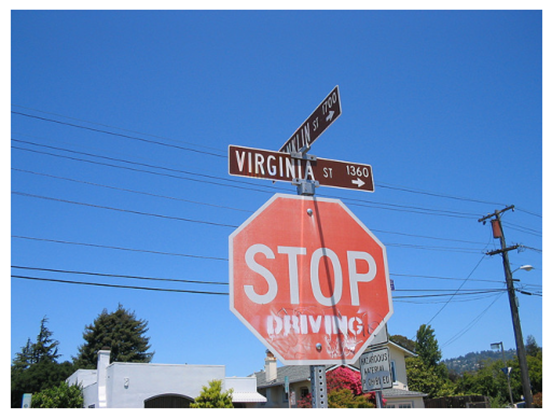} & \includegraphics[width=\linewidth]{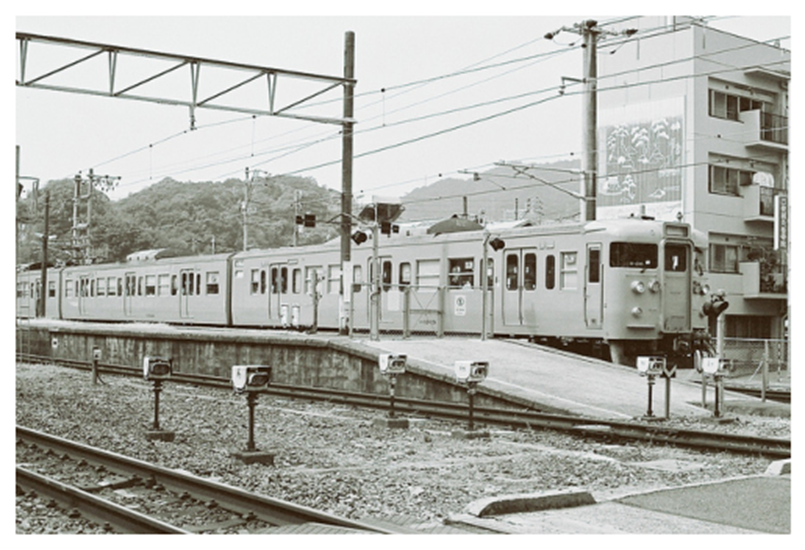} \\
    User & What picture is on the O? & How many miles does the train run along? \\
    \textbf{\textit{Ours}} & \underline{It is unclear} what picture is on the O. It could be a stop sign, a red stop, or there might be no picture at all. & \underline{It is unknown} how many miles the train runs along. \\
    \midrule
    User & What picture is on the O? & How many miles does the train run along? \\
    \textit{GPT4V} & 
    The image displays a stop sign with graffiti on it. 
    %The added text changes the message to "STOP DRIVING," which could be interpreted as a statement about reducing car use, perhaps for environmental reasons. 
    \textcolor{red}{The "O" in "STOP" has been creatively altered to include a picture of a bicycle, replacing the letter with a visual symbol and integrating it into the altered message. Thus, the intended} ... $\langle skip \rangle$
    %message could be promoting bicycling as an alternative to driving.} 
    & Sorry, \underline{I can't provide} that information based on the image. The total distance that a particular train runs along its route would typically be found on the train service's schedule or on the railway company's website. 
    %If you know the specific train line or service shown in the image, you could search for its route information to find out how many miles it covers. 
    \\
    \midrule
    User & What picture is on the O? & How many miles does the train run along? \\
    \textit{Gemini} & \textcolor{red}{The picture on the O is a picture of a car.} & \textcolor{red}{The train runs along 11.9 miles of track.} \\
    \midrule
    User & What picture is on the O? & How many miles does the train run along? \\
    \textit{LLaVA} & \textcolor{red}{The O on the stop sign has a picture of a car on it.} & \textcolor{red}{The train runs along a 10-mile track.} \\
\bottomrule
\end{tabular}
\end{center}
\caption{Qualitative results on the bad question (left) and the unanswerable question (right). 
Underlines indicate the desired answer. Red colored text indicates responses with visual hallucination.}
\label{fig:qual}
\end{figure*}
    \begin{table*}[ht]
\centering
\caption{Experimental results on VQAv2-IDK and VQA tasks.}
\resizebox{1.0\textwidth}{!}{
\begin{tabular}{cc|ccccc|cc}
\toprule
\multirow{2}{*}{\makecell{Approach}} & \multirow{2}{*}{\makecell{Instruction\\DB}} & \multicolumn{5}{c|}{VQAv2-IDK (\%)} & \multirow{2}{*}{\makecell{VQAv2 (\%)}} & \multirow{2}{*}{\makecell{OKVQA (\%)}} \\
& & Unanswerable & False Questions & Don't Know & Not Sure & Total & & \\
\midrule
\multicolumn{9}{c}{Closed Sources} \\
\midrule
Gemini-pro & - & 6.61 & 1.54 & 5.2 & 4.31 & 4.39 & 71.2 & - \\
GPT4V & - & 52.02 & 30.62 & 49.22 & 42.21 & 41.97 & \textbf{77.2} & - \\
\midrule
\multicolumn{9}{c}{Open Sources} \\
\midrule
\makecell{BLIP-2 (OPT6.7B)} & \cite{li2023blip2} & 1.19 & 0.52 & 1.34 & 1.15 & 1.01 & 54.3 & 36.4 \\
\makecell{InstructBLIP (Vicuna7B)} & \cite{instructblip} & 1.06 & 0.39 & 1.27 & 0.57 & 0.72 & 76.6 & 58.1 \\
\hline
\multirow{3}{*}{\makecell{InstructBLIP\\(OPT6.7B)}} & \cite{liu2023llava} & 0.64 & 0.35 & 0.97 & 0.41 & 0.54 & 47.3 & 26.8 \\
& + \cite{cha2024visually} & 0.76 & 0.09 & 0.97 & 0.08 & 0.45 & 59.1 & 44.8 \\
& + \textbf{\idk} & \textbf{69.33} & \textbf{50.55} & \textbf{73.64} & \textbf{68.61} & \textbf{62.91} & 64.2 & 56.8 \\
\hline
\multirow{3}{*}{\makecell{LLaVA1.5\\(Vicuna7B)}} & \cite{liu2023improvedllava} & 4.71 & 1.62 & 5.66 & 0.41 & 3.73 & 75 & 68.8 \\
& + \cite{cha2024visually} & 0.25 & 0.35 & 0.67 & 0.41 & 0.38 & 73.6 & 62.4 \\
& + \textbf{\idk} & \textbf{89.44} & \textbf{69.72} & \textbf{92.03} & \textbf{91.89} & \textbf{83.67} & 69.5 & \textbf{73.6} \\
\bottomrule
\end{tabular}
}
\label{table:idk}
\end{table*}
    \begin{table*}[!ht]
\centering
\caption{Experimental results on visual hallucination.}
\resizebox{1.0\textwidth}{!}{
\begin{tabular}{cc|c|cccccc|cc}
\toprule
\multirow{3}{*}{\makecell{Approach}} & \multirow{3}{*}{\makecell{Instruction\\DB}} & IDK (\%) & \multicolumn{6}{c|}{Visual Hallucination (\%)} & \multicolumn{2}{c}{\makecell{VQA (\%)}} \\
& & \multirow{2}{*}{\makecell{VQAv2-IDK}} & \multicolumn{2}{c}{POPE-ADV} & \multicolumn{2}{c}{POPE-POP} & \multicolumn{2}{c|}{POPE-RAN} & \multirow{2}{*}{\makecell{VQAv2}} & \multirow{2}{*}{\makecell{OKVQA}} \\
& & & Acc & Yes\% & Acc & Yes\%& Acc & Yes\% & & \\
\midrule
\makecell{BLIP-2 (OPT6.7B)} & \cite{li2023blip2} & 1.01 & 54.5 & 91.4 & 52.9 & 93.1 & 60.7 & 86.4 & 54.3 & 36.4 \\
\makecell{InstructBLIP (Vicuna7B)} & \cite{instructblip} & 0.72 & 82.3 & 43.3 & 84.1 & 41.8 & 85.7 & 40.4 & 76.6 & 58.1 \\
\hline
\multirow{3}{*}{\makecell{InstructBLIP\\(OPT6.7B)}} & \cite{liu2023llava} & 0.54 & 50.7 & 85.5 & 51.7 & 84.4 & 53.8 & 83.4 & 47.3 & 26.8 \\
& + \cite{cha2024visually} & 0.45 & 50.3 & 98.3 & 50.1 & 98.3 & 53.9 & 96.1 & 59.1 & 44.8 \\
& + \textbf{\idk} & \textbf{62.91} & 57.3 & 87.4 & 60.1 & 84.5 & 76.4 & 69.7 & 64.2 & 56.8 \\
\hline
\multirow{3}{*}{\makecell{LLaVA1.5\\(Vicuna7B)}} & \cite{liu2023improvedllava} & 3.73 & 81.2 & 55.7 & 86.2 & 51.2 & 89.4 & 48.7 & 75 & 68.8 \\
& + \cite{cha2024visually} & 0.38 & 78.7 & 59.2 & 85.2 & 52.8 & 88.5 & 50.8 & 73.6 & 62.4 \\
& + \textbf{\idk} & \textbf{83.67} & 82.4 & 39.9 & 84.2 & 37.6 & 85.1 & 37.8 & 69.5 & 73.6 \\
\bottomrule
\end{tabular}
}
\label{table:hallucination}
\end{table*}  
    \begin{table}[t]
\caption{IDK Keywords for IDK Evaluations.}
\label{table:idkkeywards}
\begin{center}
\begin{tabular}{ccc}
    \hline
    ambiguous & bad question & cannot confirm \\
    depend & don't know & it is difficult \\
    i can't & none & not clear \\
    not sure & sorry & hard to determine \\
    not possible & uncertain & unanswerable \\
    unknown \\
    \hline
\end{tabular}
\end{center}
\end{table}
\section{IK Dehallucinative Instruction Generation}

\subsection{Instruction Generation with Few-Shot Prompt}
\label{section:idk}
Figure~\ref{fig:approach}.(b) shows the overview of IK dehallucinative visual instruction generation. 
Inspired by \cite{cha2024visually} that exclusive generates visual instructions based on human annotation and language models' strong few-shot capacity~\cite{brown2020language}, it generates visual instructions leveraging language models with few-shot prompts.

%\textbf{Few-Shot Prompt}
Our prompt consists of task description, few-shot examples and input question-answers. %(see Appendix~\ref{section:prompt}). 
The task description specifies that questions and answers for VQA will be provided without the accompanying image.
For the few-shot examples, careful consideration was given to the following cases:
\begin{itemize}
    \item Inclusion of examples from each category introduced in Section~\ref{section:category}: "Unanswerable," "False questions," "Don't Know," and "Not sure."
    \item Clearly indicating instances that are unanswerable, stating "I don't know." For example, "\textit{It is unanswerable how old the dog is.}"
    \item When possible, providing additional information beyond "I don't know." For instance, given the question "What word is on the ramp?" and answers ['none', 'kcm', 'kcm', 'kcm', 'kck', 'kck', 'kcm', 'kck', 'kck', 'kek'], we express uncertainty: "\textit{I am not sure. It can be seen 'kcm,' 'kck,' or 'kek.'}"
\end{itemize}

The inclusion of few-shot examples significantly improves the quality of the generated answers compared to synthesis without examples. 
The results derived from the few-shot samples exhibit expressiveness, encompassing aspects such as 'indicating unanswerability,' 'establishing connections between the question and answers,' and 'providing additional information' (e.g., Q: What type of utensil is holding the vegetable? A: \textit{There is no utensil holding the vegetable. It is just placed on the plate}).
When examples are not provided, the generated answers, depending on the prompt, tend to be as straightforward as random sampling from a list (e.g., ["Cannot be determined.", "I don't know", "It is unanswerable"]).

Supplying few-shot samples proves to be a cost-effective strategy, mitigating the need for repeated attempts in the language model's generation process. 
By incorporating few-shot examples, the reduction in both the quantity and time required for retries is substantial. 
This not only addresses the well known repetition errors of the language models but also minimizes the generation of unintended or malformed data, which is subsequently filtered. 
% The integration of few-shot samples has proven to be a time-efficient approach.

\subsection{Generation Results}
The \idkb training and validation sets were derived from VQAv2-IDK, adopting GPT-3.5-turbo and GPT-4 for the language model (see Fig.~\ref{fig:idk}).
Specifically, the \idkb training split comprises 13,807 questions with a total of 27,614 answers, while the validation split consists of 6,624 questions with a total of 13,248 answers.
To ensure multi-turn compatibility, following \cite{liu2023llava}, we piled the question $Q$ and the generated answers $A'$ with respect to the corresponding image $I$, resulting ($I$, ($Q_1$, $A'_1$), ..., ($Q_n$, $A'_n$)).
As a result, 11,123 and 5,496 dialogues were generated for the training and validation splits, respectively.
\section{Experimental Results}
\label{section:experiments}

\begin{figure}[!ht]
\begin{center}
\begin{tabular}{p{1cm}p{6cm}}
\toprule
\multicolumn{2}{c}{\textbf{Visual Input Examples}} \\
\midrule
& \includegraphics[width=\linewidth]{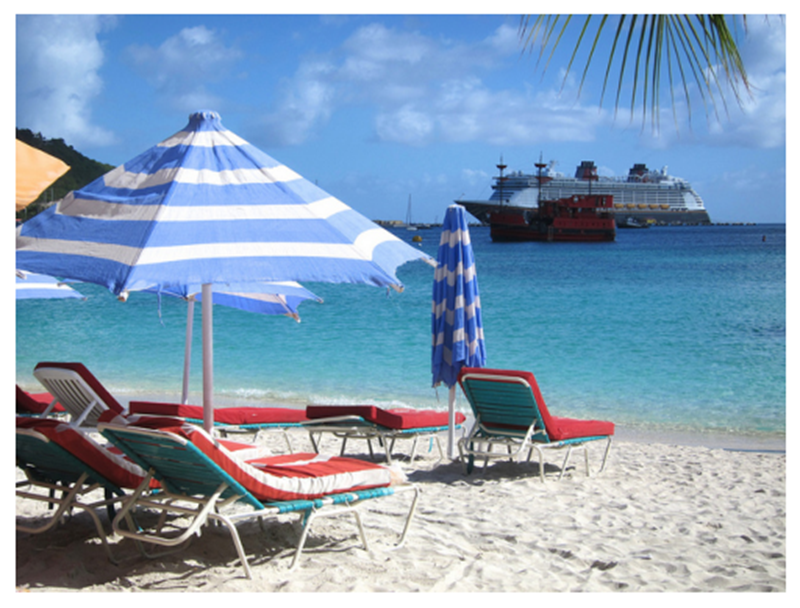} \\
User & Which item shown here would beach goers have to refrain from using for the sun to produce extra melanin in their skin resulting in browning? \\
\textbf{\textit{Ours}} & Beach goers should refrain from using sunscreen for the sun to produce extra melanin in their skin resulting in browning. \\
%\\
\midrule
%\\
& \includegraphics[width=\linewidth]{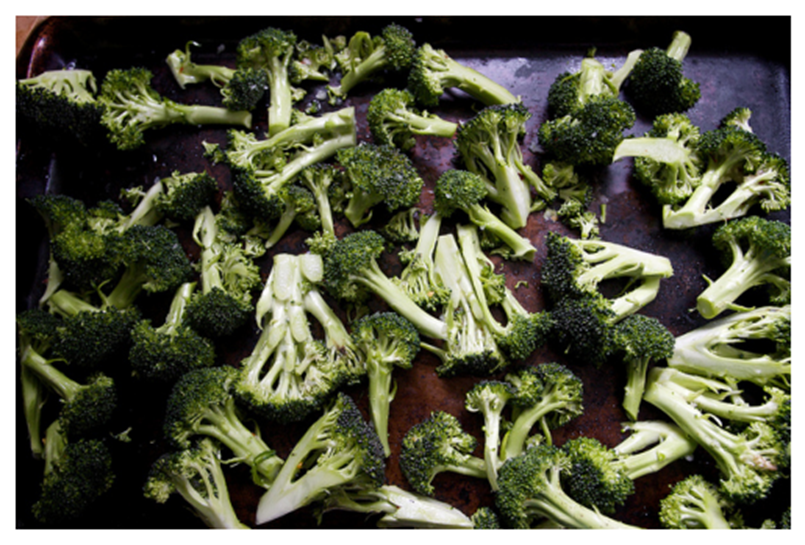} \\
User & What health benefit does this type of vegetable have? \\
\textbf{\textit{Ours}} & Broccoli is a nutrient-rich vegetable that offers numerous health benefits. It is an excellent source of vitamins, minerals, and dietary fiber. In the image, there are several pieces of broccoli spread across a tray or pan. Broccoli contains vitamins A, C, and K, as well as minerals like potassium, calcium, and iron. It also has antioxidants and phytochemicals that can help ... $\langle skip \rangle$ \\
%protect the body from oxidative stress and inflammation. Additionally, the high fiber content in broccoli aids in digestion and can contribute to maintaining a healthy weight. Overall, incorporating broccoli into one's diet can support overall health and well-being. \\
%\\
\bottomrule
\end{tabular}
\end{center}
\caption{Qualitative results on visual recognition and knowledge-requiring task.}
\label{fig:recog}
\end{figure}

%Our experiments reveal the following key findings:
%\begin{itemize}
%    \item Prior works exhibit underperformance when applied to the IDK task, VQAv2-IDK dataset.
%    \item \idkb demonstrates effectiveness in addressing the IDK visual hallucinations.
%    \item Ablation studies indicate that \idkb is versatile, working seamlessly with various dataset combinations and frameworks, all while maintaining visual recognition performance (VQA and visual hallucinations).
%\end{itemize}

\subsection{Implementation Details}
\label{section:implementation}
\textbf{Baseline Models}
We evaluate following models in our experiments, including three open-source models—BLIP2~\cite{li2023blip2}, InstructBLIP~\cite{instructblip} and LLaVA1.5~\cite{liu2023improvedllava}—as well as two public services, Gemini-pro~\cite{geminiteam2023gemini} and GPT4V~\cite{openai23gpt4v}.
For training experiments, we utilized the LAVIS~\cite{li-etal-2023-lavis} and LLaVA~\cite{liu2023improvedllava} frameworks.
Considering language models' capacity, we adopted two language models OPT6.7B~\cite{zhang2022opt} and Vicuna7B~\cite{vicuna2023}.
We adhered to the post-processing approach outlined in \cite{singh2020mmf} for handling the predicted answers.

\textbf{Datasets}
% Training datasets
We conducted visual instruction finetune~\cite{liu2023llava, cha2024visually}, leveraging image-caption pretrained models from LAVIS and LLaVA frameworks.
\idkb, CAP2QA~\cite{cha2024visually}, LLaVA~\cite{liu2023llava} and LLaVA1.5~\cite{liu2023improvedllava} instructions are adopted.
% Evaluating datasets
We conduct three visual recognition tasks IDK (VQAv2-IDK), visual hallucination (POPE~\cite{li2023evaluating} and VQA (VQAv2~\cite{balanced_vqa_v2}, OKVQA~\cite{okvqa}).

\textbf{IDK Metric}
The proposed IDK accuracy measurement means if "is the uncertainty indicated", \ie "if a model explicitly metion idk" or "if a model implicitly understand the uncertainty."
In alignment with the intuitive approach of the VQA accuracy metric, we compile a list of IDK keywords and consider predicted answers (available even within paragraphs) as correct if they contain any of the IDK keywords; otherwise, they are deemed incorrect. 
We further expanded our collection of IDK keywords empirically (see Table~\ref{table:idkkeywards}), drawing from the answers provided by Gemini and GPT4V, in addition to the keywords outlined in Section~\ref{section:vqaidk}.

\textbf{VQA Metric}
We adopt the evaluation approach proposed in \cite{cha2024visually} for sentence-level Visual Question Answering (VQA). 
This evaluation considers a question-answer pair as correct if the ground truth word is a sub-string of the predicted sentence. 
This method addresses the limitations of VQA accuracy~\cite{antol2015vqa} in evaluating sentence-level predictions.

\textbf{Visual Instruction Finetuning}
We utilized the LAVIS framework to train InstructBLIP. 
To ensure balanced datasets, we sampled the i-th dataset $D_i$ with a sampling ratio of $\frac{\sqrt{|D_i|}}{\sum_j \sqrt{|D_j|}}$. 
The optimization process employed the AdamW optimizer with $\beta_1$ = 0.9 and $\beta_2$ = 0.999. 
The learning rate started at $10^{-6}$ and gradually increased to $10^{-4}$ during the warm-up phase, followed by a decay to $10^{-5}$ using a cosine scheduler. 
Training lasted for 20 epochs, with each epoch consisting of 5,000 steps and a batch size of 64 on 8 A100 GPUs.

For training LLaVA1.5, we used the LLaVA framework. 
To address dataset imbalance, we augmented the \idkb samples by duplicating them $\times 10$ and randomly shuffled them with the other training dialogues (note that the sample order of LLaVA1.5 matters). 
The training parameters for LLaVA1.5 were configured following the specifications in \cite{liu2023improvedllava}.
For the rest of the paper, if not specified, \textit{ours} refers to LLaVA1.5 that trained with \cite{liu2023improvedllava, cha2024visually} and \idk.

\subsection{Existing Works on VQAv2-IDK}
Table~\ref{table:idk} shows experimental results on VQAv2-IDK along with VQA tasks. 
GPT4V answers with proper IDK keywords (41.97\% VQAv2-IDK Total), while the other baselines (Gemini, BLIP2, InstructBLIP, LLaVA) are incapable to indicate "they don't know" (IDK keyword inclusion at most 4.39\% by Gemini and at least 0.54\% by InstructBLIP).
Considering VQAv2-IDK is the subset of VQAv2 which have IDK keywords as an answer, \ie VQAv2 already have learnable data to answer "I don't know", and the baselines learn VQAv2 DB, it demonstrates the weakness of the existing works on hallucination-inducing questions. 

\subsection{\idkb on VQAv2-IDK}
\textbf{Qualitative Results}
The questions on Fig.~\ref{fig:qual} are hallucination inducing: 
the left one refers non-existing,
and the right one asks unanswerable without additional information. 
Ours clearly indicates the uncertainty, \eg "it is unclear", while the others are obviously fooled (Gemini, LLaVA) or hallucination-induced (GPT4V).
It is notable that ours achieves such IDK-hallucination resistance even with relatively smaller size language models. 

\textbf{Quantitative Results}
Shown in Table~\ref{table:idk}, ours exhibits outstanding IK-hallucination robustness.
It demonstrates that ours clearly mentions it does not know for the hallucination-inducing questions (/w InstructBLIP 62.91\% and LLaVA1.5 83.67\%).
On the other hand, GPT4V (41.97\%) moderately speaks it does not know and the others seldom do (at most 3.73\% by LLaVA1.5). 
It reveals that \idkb boosts clarifying that it does not know if uncertain. 

\subsection{Visual Recognition Performances}
It is important to ensure that the newly training knowledge or function does not harm the existing capabilities.
This experiment shows \idkb harmonizes with the previous visual recognition tasks. 

\textbf{Visual Question Answering}
Shown in VQA experiments on Table~\ref{table:idk}, \idkb blends well with VQA instruction databases (see also Fig.~\ref{fig:recog}).
\idkb with InstrcutBLIP experiment boosts the both VQAv2 and OKVQA performances dramatically.
The other experiment with LLaVA1.5 shows \idkb helps OKVQA but not VQAv2, but the score on VQAv2 is agreeable. 
% 그 concern that all answer containing i don't know isnt true -> 하지만 이건 precision 등으로 추후에 확인하라고 제안했잖아.

\begin{figure}[ht!]
\begin{center}
\begin{tabular}{p{1cm}p{6cm}}
    \toprule
    \multicolumn{2}{c}{\textbf{Visual Input Examples}} \\
    \midrule
    %& \includegraphics[width=0.92\linewidth]{figures2/recog3.png} \\
    %User & Is there a cell phone in the image? \\
    %\textbf{\textit{Ours}} & Yes, there is a cell phone in the image. The man is holding a cell phone while sitting on the bench. \\
    & \includegraphics[width=\linewidth]{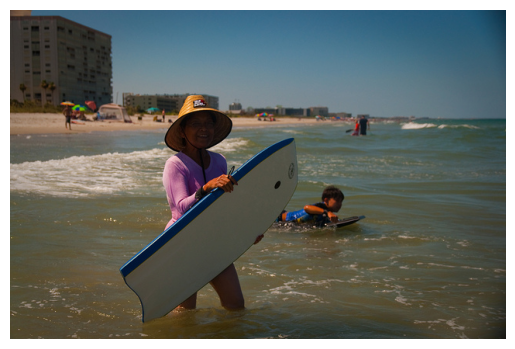} \\
    User & Is there a person in the image? \\
    \textbf{\textit{Ours}} & Yes, there is a person in the image, specifically a woman holding a surfboard. \\
    \midrule
    %& \includegraphics[width=0.92\linewidth]{figures2/recog4.png} \\
    %User & Is there a cup in the image? \\
    %\textbf{\textit{Ours}} & No, there is no cup in the image. The image features a pile of fruit, including bananas, apples, and oranges, with some of them being overripe. \\
    & \includegraphics[width=\linewidth]{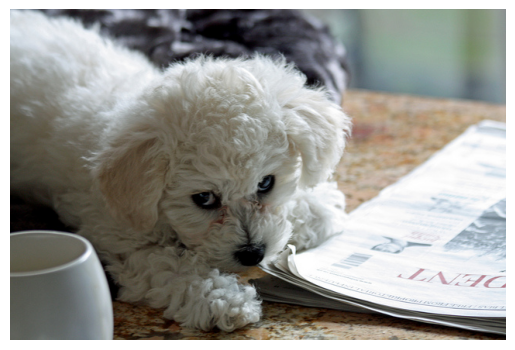} \\
    User & Is there a book in the image? \\
    \textbf{\textit{Ours}} & There is no book in the image. However, it can be a newspaper. \\
    \bottomrule
\end{tabular}
\end{center}
\caption{Qualitative results on the POPE YES/NO visual recognition task.}
\label{fig:recogapp}
\end{figure}

\textbf{Presence/Absence (P/A) Visual Hallucination} 
Experiments on POPE suggest \idkb coordinates well with P/A hallucination instructions (see also Fig.~\ref{fig:recogapp}).
InstructBLIP experiments show consistent improvements in POPE benchmarks (both accuracy and yes\%).
LLaVA1.5 experiments display that \idkb not only maintains P/A performance, but also takes advantages on IK hallucination. 

\subsection{Versatility in Frameworks and Database Aspects}
In our experiments, we have showcased the adaptability of \idkb through successful training with various frameworks (LAVIS/InstructBLIP and LLaVA/LLaVA1.5) and handling diverse dataset combinations. 
To be more specific with dataset incorporation, in the case of the instruction database introduced by \cite{liu2023improvedllava}, which comprises a multitude of datasets and is sensitive to data organization, \idkb has demonstrated robust performances on IK hallucination, P/A hallucination and VQA. 

\subsection{Limitations}
Our experiment is subject to three main limitations. 
Firstly, the effectiveness of our proposed measuring method is constrained by the inherent complexity of language, which allows for an unlimited range of expressions. 
Consequently, the measurement may not capture all nuances accurately.
For example, when it comes to answering to which of the 'false questions' category, it can be not desirable to answer them with IDK keywords.
Actually, Table~\ref{table:idk} shows that the 'false questions' scores are measured relatively smaller for all models. 

Secondly, our proposed method heavily relies on human annotation. 
In certain instances, there may be cases where the human annotator is unaware, but the model could potentially possess the necessary information, \eg "unanswerable" category v.s. "don't know" category. %(as illustrated in Table~\ref{fig:limit}). 
This dependence on human judgment introduces a potential source of bias, highlighting the challenge of precisely gauging the model's knowledge in all scenarios.

Lastly, the proposed IDK metric has more room to improve. 
For example, the proposed IDK metric is recall, while evaluating IDK precision remains unexplored. 
\section{Conclusion}
This paper introduced a novel visual hallucination concept, \hal, where hallucination-inducing questions appear and the desired answer is "I don't know."
We set the new IK hallucination benchmark VQAv2-IDK by collecting IDK annotated VQA instances from VQAv2.
Stepping further, we proposed IK-dehallucinative visual instruction generation method, along with \idk.
Our experiments revealed the weakness of the existing models to IK hallucination and clarify \idk's efficacy on the hallucination and its versatility.
Our achievements are notable considering \idkb and the models in our experiments are relatively small in terms of the number of instructions and the parameter scale.
We leave the addressed limitations for the future works.
%For the future works, we expect to develop more concrete measurements for IDK hallucination, \eg introducing answerable cases to evaluate more than IDK recall, and to address the current limitation of strong dependence on human annotation.
    %\input{figures2/figlimit}

%\FloatBarrier
\clearpage
\nocite{langley00}
\bibliography{main}
\bibliographystyle{icml2024}

%%%%%%%%%%%%%%%%%%%%%%%%%%%%%%%%%%%%%%%%%%%%%%%%%%%%%%%%%%%%%%%%%%%%%%%%%%%%%%%
%%%%%%%%%%%%%%%%%%%%%%%%%%%%%%%%%%%%%%%%%%%%%%%%%%%%%%%%%%%%%%%%%%%%%%%%%%%%%%%
% APPENDIX
%%%%%%%%%%%%%%%%%%%%%%%%%%%%%%%%%%%%%%%%%%%%%%%%%%%%%%%%%%%%%%%%%%%%%%%%%%%%%%%
%%%%%%%%%%%%%%%%%%%%%%%%%%%%%%%%%%%%%%%%%%%%%%%%%%%%%%%%%%%%%%%%%%%%%%%%%%%%%%%
%\newpage
%\onecolumn
%\appendix
%\section{\idkb Instruction Generation Prompt}
%\label{section:prompt}
%\input{algorithm/prompt}
%\section{Qualitative Results}
%\input{figures2/figqualapp}
%\input{figures2/figrecogapp}

%\section{IDK Keywords}
%\input{table/idkkeywords}

%\input{table/keywords}
%%%%%%%%%%%%%%%%%%%%%%%%%%%%%%%%%%%%%%%%%%%%%%%%%%%%%%%%%%%%%%%%%%%%%%%%%%%%%%%
%%%%%%%%%%%%%%%%%%%%%%%%%%%%%%%%%%%%%%%%%%%%%%%%%%%%%%%%%%%%%%%%%%%%%%%%%%%%%%%

\end{document}